\documentclass[sigconf]{aamas} 

\usepackage{balance} 
\usepackage{hyperref}
\usepackage{graphicx}
\usepackage{amsmath}
\usepackage{algorithm}
\usepackage{multirow}
\usepackage{algorithmic}
\MakeRobust{\ref}

\makeatletter

\setcopyright{ifaamas}
\acmConference[AAMAS '22]{Proc.\@ of the 21st International Conference
on Autonomous Agents and Multiagent Systems (AAMAS 2022)}{May 9--13, 2022}
{Online}{P.~Faliszewski, V.~Mascardi, C.~Pelachaud,
M.E.~Taylor (eds.)}
\copyrightyear{2022}
\acmYear{2022}
\acmDOI{}
\acmPrice{}
\acmISBN{}

\acmSubmissionID{719}

\title{ACuTE: Automatic Curriculum Transfer from \\Simple to Complex Environments}

\author{Yash Shukla}
\affiliation{
  \institution{Tufts University}
  \city{Medford}
  \country{USA}}
\email{yash.shukla@tufts.edu}

\author{Christopher Thierauf}
\affiliation{
  \institution{Tufts University}
  \city{Medford}
  \country{USA}}
\email{christopher.thierauf@tufts.edu}

\author{Ramtin Hosseini}
\affiliation{
  \institution{Tufts University}
  \city{Medford}
  \country{USA}}
\email{ramtin.hosseini@tufts.edu}

\author{Gyan Tatiya}
\affiliation{
  \institution{Tufts University}
  \city{Medford}
  \country{USA}}
\email{gyan.tatiya@tufts.edu}

\author{Jivko Sinapov}
\affiliation{
  \institution{Tufts University}
  \city{Medford}
  \country{USA}}
\email{jivko.sinapov@tufts.edu}

\begin{abstract}
Despite recent advances in Reinforcement Learning (RL), many problems, especially real-world tasks, remain prohibitively expensive to learn. To address this issue, several  lines  of  research  have  explored  how  tasks,  or  data  samples themselves, can be sequenced into a curriculum to learn a problem that may otherwise be too difficult to learn from scratch. However, generating and optimizing a curriculum in a realistic scenario still requires extensive interactions with the environment. To address this challenge, we formulate the {\it curriculum transfer} problem, in which the schema of a curriculum optimized in a simpler, easy-to-solve environment (e.g., a grid world) is transferred to a complex, realistic scenario (e.g., a physics-based robotics simulation or the real world). We present ``ACuTE'', Automatic Curriculum Transfer from Simple to Complex Environments, a novel framework to solve this problem, and evaluate our proposed method by comparing it to other baseline approaches (e.g., domain adaptation) designed to speed up learning. We observe that our approach produces improved jumpstart and time-to-threshold performance even when adding task elements that further increase the difficulty of the realistic scenario. Finally, we demonstrate that our approach is independent of the learning algorithm used for curriculum generation, and is Sim2Real transferable to a real world scenario using a physical robot.
\end{abstract}


\newcommand{\BibTeX}{\rm B\kern-.05em{\sc i\kern-.025em b}\kern-.08em\TeX}
\newcommand{\labeltext}[2]{%
  \@bsphack
  \csname phantomsection\endcsname 
  \def\@currentlabel{#1}{\label{#2}}%
  \@esphack
}

\begin{document}


\pagestyle{fancy}
\fancyhead{}


\maketitle 


\section{Introduction}

Curriculum Learning (CL) attempts to optimize the order in which an agent accumulates experience, increasing performance while reducing training time for complex tasks~\cite{narvekar2020curriculum,foglino2019optimization,narvekar2017autonomous}. The core of CL is to generalize the experience and knowledge acquired in simple tasks and leverage it to learn complex tasks. Viable results are achieved in simulation, where the system dynamics can be easily modeled and the environment is predictable. One major limitation of many curriculum learning approaches is that the time to generate the curriculum is greater than the time to learn the target task from scratch, which prohibits the use of such methods in complex, real-world, high-fidelity domains \cite{narvekar2020curriculum}. The useful scenario of transfer to the real world remains challenging: System identification, domain adaptation, and domain randomization have performed Sim2Real transfer by attempting to match simulation with the physical environment (see~\ref{sim2real}), but these methods are elaborate and time-consuming if the simulation dynamics are expensive.

Since the dynamics of high-fidelity (HF) environments may not lend themselves to optimize a curriculum \cite{zhang2020automatic, openai2020neurips}, we propose learning the curriculum in a simplified version of the HF environment, which we call the low-fidelity (LF) environment. Parameters from each LF task can then be transferred, generating a corresponding task in the HF environment. This curriculum transfer problem is an open question (see ~\cite{narvekar2020curriculum}), and to our knowledge, our novel approach is the first to address this problem. We refer to this as transferring the {\it schema} of the curriculum: only task parameters are transferred as to address situations where policies and value functions cannot be directly transferred due to differences in the observation and action spaces. High-level task descriptions can be used to model inter-task relationships, and tasks with similar task-descriptors are shown to aid positive transfer~\cite{rostami2020using, sinapov2015learning}.
We show that our curriculum transfer approach leads to a quicker convergence even in cases where the dynamics of the LF and HF environments are different enough such that traditional domain adaptation methods do not produce a sufficient boost in learning. 
\begin{figure*}
\centering
    \includegraphics[width=0.7\linewidth, height=0.3\linewidth]{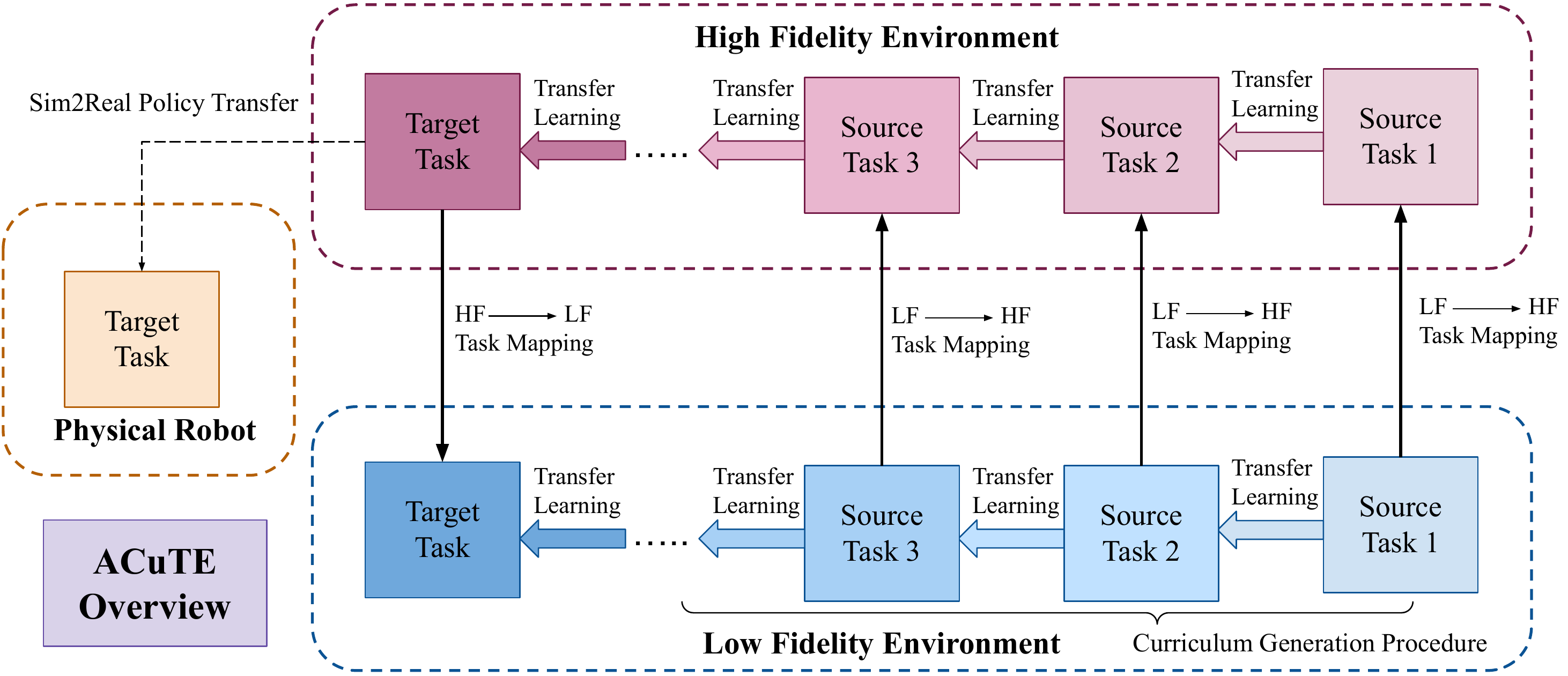}
    \vspace{-1em}
    \caption{Overview of the proposed curriculum transfer approach ACuTE. The curriculum is generated and optimized in the low-fidelity environment, which is then mapped to generate a curriculum of the high-fidelity environment and learned before learning the final target task. The policy is then transferred to a physical robot.}
    \label{fig:overview}
    \vspace{-1em}
\end{figure*}

An overview of our approach is shown in Fig~\ref{fig:overview}. We consider a complex task and call it the HF target task, and map it to its simplified LF representation. The simplified dynamics of the LF environment allows curricula generation and experimentation to avoid costly setup and expensive data collection associated with the HF environment. Once an optimized curriculum is generated in the LF environment, the task parameters are mapped to obtain their respective HF counterparts. We learn these source tasks iteratively, transferring skills, before learning the target task. Finally, we perform a demonstration with Sim2Real transfer from the HF environment to test our effectiveness on a physical TurtleBot.

In this work, we perform extensive experimental evaluation to demonstrate that curriculum transfer enables the agent to reduce the overall target task time compared to baselines. Through ACuTE, we propose an autonomous curriculum transfer method, which we refer to as ``Automated Curriculum Transfer'', that parameterizes the target task to generate and optimize the sequence of source tasks. We notice quick and efficient learning compared to baseline approaches present in literature such as Domain Adaptation \cite{carr2019domain}, Self-Play \cite{sukhbaatar2018intrinsic} and Teacher-Student curriculum learning~\cite{DBLP:journals/tnn/MatiisenOCS20}. Additionally, we observe an improved jumpstart and time to threshold performance even when we add elements that make the HF target task too difficult to learn without a curriculum. We demonstrate that our approach is independent of the learning algorithm by showing improved performance in the HF environment when using a different learning algorithm from the one used when optimizing the curriculum, and also demonstrate positive transfer with imperfect mapping between the two environments. We observe that the Sim2Real transfer achieves successful task completion performance, equivalent to the HF agent's performance, on a physical TurtleBot.
\section{Related Work}
\label{sec:related_work}

\textbf{Transfer Learning} uses knowledge from learned tasks and transfers it to a complex target task~\cite{taylor2009transfer}. Policy transfer is one such approach, in which the policy learned in a source task is used to initialize the policy for the next task in the curriculum \cite{narayan2017seapot,da2019survey, taylor2009transfer, lazaric2012transfer}. One popular transfer learning technique is to transfer the value function parameters learned in one task to initialize the value function of the next task in curriculum \cite{liu2006value, abel2018policy, liu2019value, lazaric2012transfer}.  

\textbf{Sim2Real Transfer}\labeltext{Sim2Real Transfer}{sim2real} allows a model to be trained in simulation before deploying onto hardware, reducing time, cost, and safety issues. However, it encounters what Tobin~\emph{et.~al.}~\cite{tobin2017domain} describe as the ``Reality Gap'' where a model does not allow an agent to train on realistic details. The same authors introduce ``domain randomization'' as a solution, later expanded upon by Peng~\emph{et.~al.} \cite{peng2018sim}. Continual learning on incremental simulations can help tackle the sample inefficiency problem of domain randomization~\cite{josifovski2020}.
In contrast, Operational Space Control \cite{kaspar2020sim2real} avoids domain randomization while speeding up training with fewer hyperparameters.
Carr~\emph{et.~al.} \cite{carr2019domain} proposed a domain adaptation strategy approach, where state knowledge is transferred across semantically related games.
Unlike aforementioned works, in this paper, we transfer the curriculum and not the policy to handle situations where Sim2Real fails,~e.g.,~when the observation and action spaces of the simulation and real environment are different.

\textbf{Curriculum Learning} was introduced in the early 1990's, where it was used for grammar learning \cite{elman1993learning}, control problems \cite{sanger1994neural} and in supervised classification tasks \cite{bengio2009curriculum}. CL has been extensively used for RL, with applications ranging from complex games \cite{gao2021efficiently,wu2018master}, to robotics \cite{karpathy2012curriculum}, and self-driving cars \cite{qiao2018automatically}.  In \cite{narvekar2020curriculum}, the authors propose a framework for
curriculum learning (CL) in RL, and use it to classify and survey existing CL algorithms. The three main elements of CL are task generation, task sequencing, and transfer learning. Task generation produces a set of source tasks that can be learned before learning the target task \cite{silva2018object,kurkcu2020autonomous}.
Task sequencing orders the generated tasks to facilitate efficient transfer from the source to the target task \cite{narvekar2017autonomous, sukhbaatar2018intrinsic, DBLP:journals/tnn/MatiisenOCS20}. Metaheuristic search methods are a popular tool to evaluate the performance of the task sequencing optimization framework \cite{foglino2019optimization}. In our work, we propose a framework to generate the source tasks and optimize their sequence, while evaluating performance against three baseline approaches. In most existing methods, generating and optimizing the curricula to learn a complex task is still time-consuming and sometimes takes longer compared to learning from scratch. Our proposed framework addresses this concern by generating, optimizing, and then transferring the schema of the curriculum from a simple and easy-to-learn environment to a complex and realistic environment.

\vspace{-0.75em}

\section{Theoretical Framework}

\subsection{Markov Decision Processes}

An episodic Markov Decision Process (MDP) $M$ is defined as a tuple $(\mathcal{S},\! \mathcal{A},\! p,\! r,\! \gamma)$, where $\!\mathcal{S}\!$ is the set of states, $\!\mathcal{A}\!$ is the set of actions, $\!p(s'\!|s,a)\!$ is the transition function,~   
$\!r(s'\!,a,s)\!$ is the reward function and $\gamma\!\in\! [0,1]$ is the discount factor. For each timestep $t$, the agent observes a state $s$ and performs an action $a$ given by its policy function $\pi_\theta(a|s)$, with parameters $\theta$. The agent's goal is to learn an \emph{optimal policy $\pi^*\!$}, maximizing its discounted return $G_0 = \sum^{K}_{k = 0}\!\gamma^k\! r(s'_k,a_k,s_k) $ until the end of the episode at timestep $K$.

\begin{figure*}
    \begin{center}
        \begin{tabular}{ c c c }
            {\includegraphics[width=0.29\textwidth, height=0.215\textwidth]{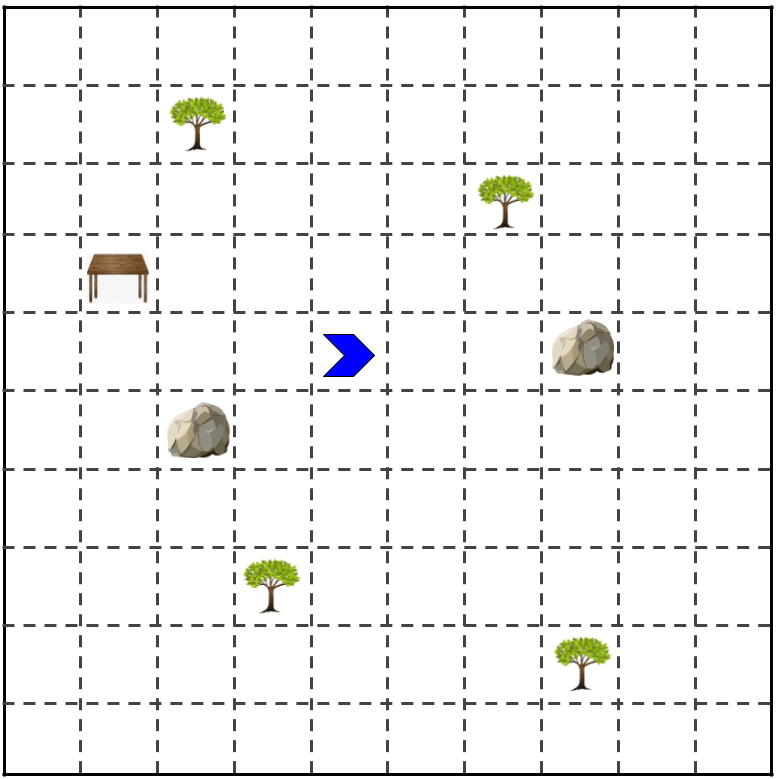}} &
            {\includegraphics[width=0.29\textwidth]{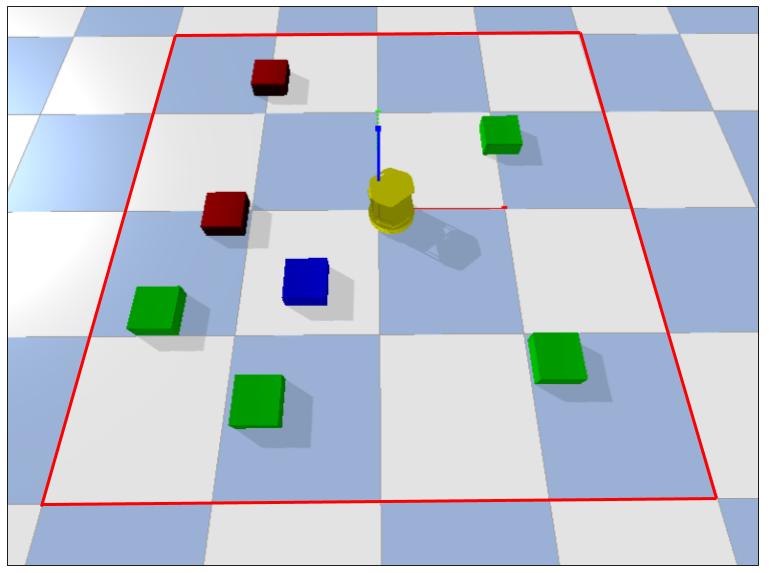}}&
            {\includegraphics[width=0.29\textwidth]{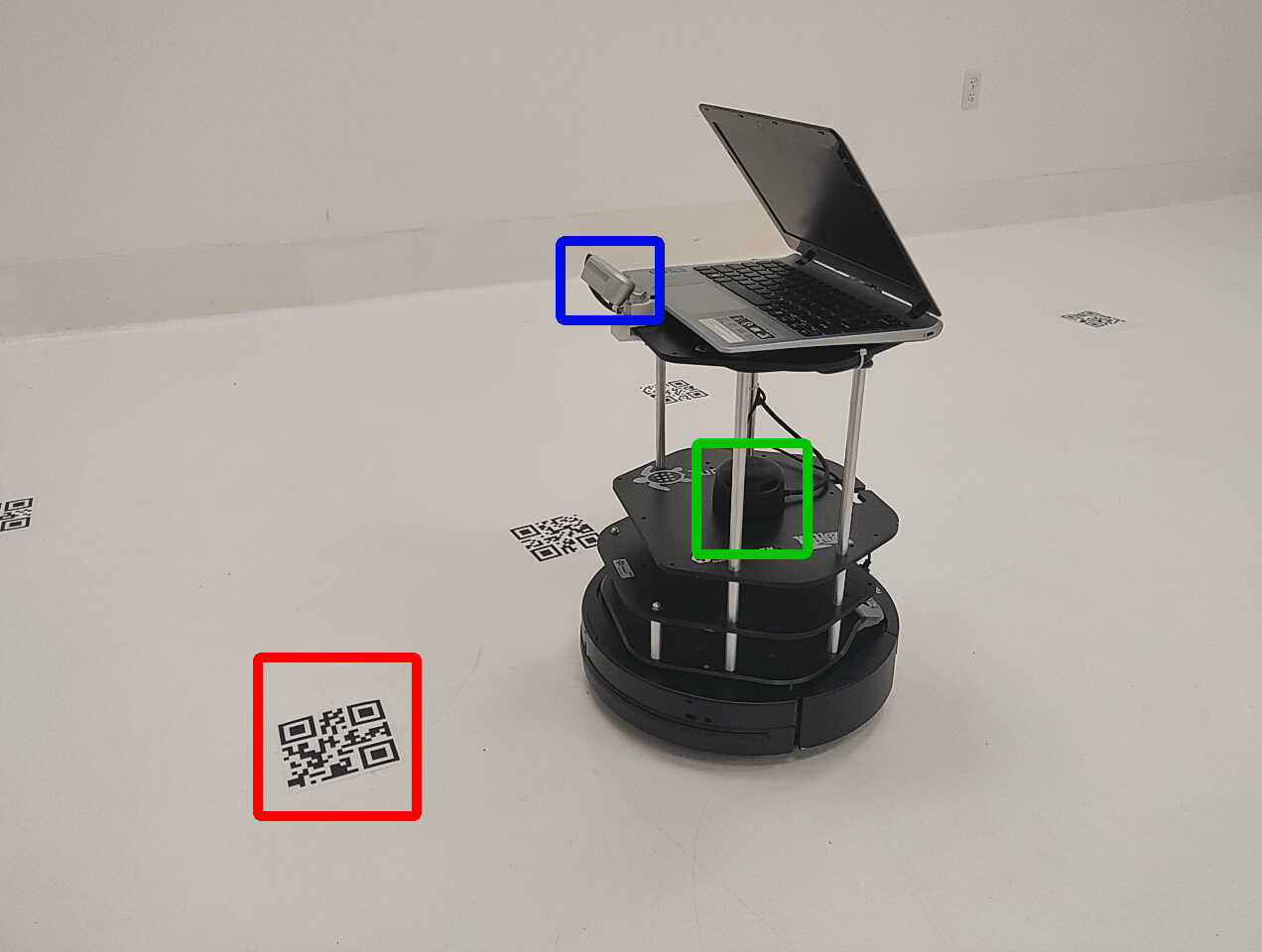}}
            \\
            {\small{\textbf{(a) Target task in Low Fidelity Environment}}} &
            {\small{\textbf{(b) Target task in High Fidelity Environment}}} &
            {\parbox{0.29\textwidth}{\small{\textbf{(c) Target task in Physical Environment, using a camera (blue) to interact with fiducials (red). LIDAR (green) is also visible.}}}}
        \end{tabular}
    \end{center}
    \vspace{-1em}
    \caption{Illustration of the final target task in LF, HF and physical  environment. The agent performs navigation, breaking action on 2 trees and 1 rock and then crafts a stone-axe at the crafting table to successfully terminate the task. \label{fig:Final_tasks}}
\end{figure*}
\vspace{-1em}

\vspace{-0.5em}
\subsection{Curriculum Learning (CL)} \label{section:CL}

We define a task-level curriculum as:  \\
\textit{Let $\mathcal{T}$ be a set of tasks, where $M_i = (\mathcal{S}_i, \mathcal{A}_i, p_i, r_i)$ is a task in $\mathcal{T}$. Let $\mathcal{D^T}$ be the set of all possible transition samples from tasks in $\mathcal{T}$ : $\mathcal{D^T} = \{(s,a,r,s') \, | \, \exists \, m_i \!\in \!\mathcal{T} \: s.t. \: s\! \in\! \mathcal{S}_i, \: a \!\in\! \mathcal{A}_i, \: s' 
\! \sim\! p_i(\cdot|s,a),\: r \leftarrow r_i(s,a,s') \}$. A curriculum $C = [M_1, M_2, \ldots, M_n]$ is an ordered list of tasks, where $M_i$ is the $i^{th}$ task in the curriculum. The ordered list signifies that samples from $M_i$ must be used for training a policy before samples from $M_{i+1}$ are used. The sequence of tasks terminates on the target task $M_n$.}

\vspace{-0.5em}
\subsection{Problem Formulation}

The aim of CL is to generate a curriculum and train an agent on a sequence of tasks $\{M_1, M_2, ..., M_U\}$, such that the agent's performance on the final target task $(M_U)$ improves relative to learning from scratch. The domains $\mathcal{T}^{HF}$ and $\mathcal{T}^{LF}$ of possible tasks are sets of MDPs in the high-fidelity (HF) and the low-fidelity (LF) environments, respectively. An individual task can be realized by varying a set of \textit{parametric variables} and subjecting the task to a set of \textit{constraints}. 
The \textit{parametric variables $P$} are a set of attribute-value pair features $[ P_1,..., P_n ]$ that parameterize the environment to produce a specific task. Each $P_i \in P$ has a range of possible values that the feature can take while the $constraints$ of a domain are a set of tasks attained by determining the goal condition $P_G$.

Let $\mathcal{C}_U^{\mathcal{T}_{LF}}$ be the set of all curricula over tasks $\mathcal{T}^{LF}$ of length $U$ in the LF environment. Similarly, let $\mathcal{C}_U^{\mathcal{T}_{HF}}$ be the set of all curricula over tasks $\mathcal{T}^{HF}$ of length $U$ in the HF environment. 
The goal is to find a curriculum ${c_U}^{\mathcal{T}_{LF}}$ in the LF environment that can be transferred through a set of mapping functions $\mathcal{F} := \{f_1, f_2, \ldots, f_n \}$ to attain the curriculum in the HF environment ${c_U}^{\mathcal{T}_{HF}}$. A mapping function maps the parametric variables in the LF environment ($P^{LF}$) to the parametric variables in the HF environment ($P^{HF}$). We characterize the mapping as an affine transformation given by:
\[P^{HF} = A\odot P^{LF} + B \]
where $A = [a_1, \ldots, a_n]^T \in \mathbb{R}^n$ and $B = [b_1, \ldots, b_n]^T \in \mathbb{R}^n$ denote linear mapping and translation vector and $\odot$ is the Hadamard product. Thus, a parameter mapping ($f_i : P^{LF}_i \rightarrow{} P^{HF}_i$) is given by:
\[P^{HF}_i = a_i P^{LF}_i + b_i\]
The source tasks of the curriculum in the HF environment are learned before final target task as described in Section~\ref{section:CL}.

\vspace{-0.5em}
\subsection{Running Example}

For the physical environment shown in Fig~\ref{fig:Final_tasks}c, we generate Crafter-TurtleBot (Fig~\ref{fig:Final_tasks}b), a realistic simulation of the physical environment. The aim is to learn a policy in this high-fidelity (HF) environment, through an automated curriculum transfer from the low-fidelity (LF) environment (Fig~\ref{fig:Final_tasks}a), and perform  Sim2Real Policy transfer from the HF environment to execute the task in the physical environment.

The agent's goal is to break 2 trees to collect 2 pieces of wood, break a rock to collect a stone and craft a stone axe at the crafting table. The agent needs to navigate, face the object and perform the break action to collect it in inventory. The parametric variables for this task are the width $(P_W)$ and height $(P_H)$ of the navigable area, the number of trees $(P_{T,e})$, rocks $(P_{R,e})$, and crafting table $(P_{CT})$ present, the number of wood $(P_{T,i})$ and stones $(P_{R,i})$ present in the inventory of the agent when the episode starts, and the goal $(P_G)$ of the task. The goal is drawn from a discrete set, which can be navigating to an item, breaking a subset of the items present in the environment or crafting the stone axe.

As described in Fig~\ref{fig:overview}, the HF target task is mapped to its LF equivalent. The simplified LF dynamics allow efficient curriculum optimization. Once the curriculum is generated, each task of the LF is mapped back to generate an equivalent HF task, which are learned through a curriculum to develop a successful task policy for the target HF task. This policy is then transferred to the Physical Robot through a Sim2Real Transfer.


\begin{figure*}
\centering
    \includegraphics[width=0.78\linewidth, height=0.25\linewidth]{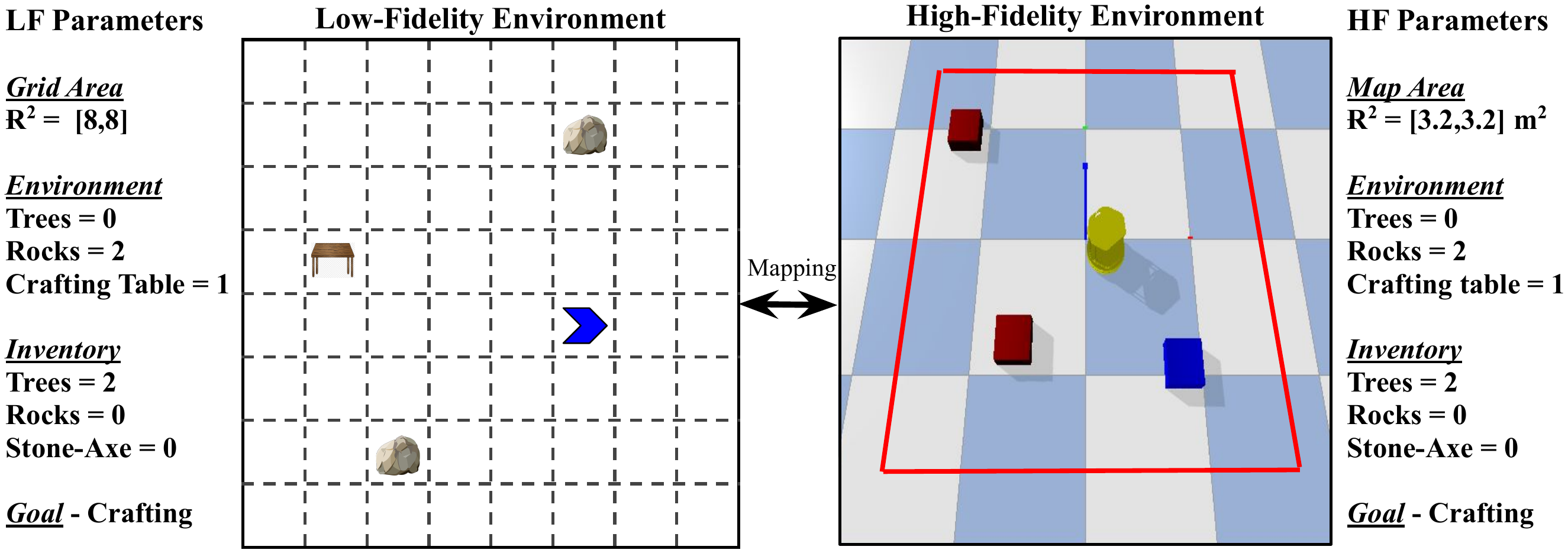}
    \vspace{-1em}
    \caption{\small{Illustrative example of the low-fidelity to high-fidelity mapping in the crafting task.}}
    \label{fig:one-to-one_example}
\end{figure*}

\vspace{-0.5em}
\subsection{Curriculum Transfer Approach}
\label{section:Algorithm for curriculum transfer}
ACuTE consists of three parts: Generating a LF target task, Curriculum generation in LF, and Task sequencing and learning in HF. Algorithm~\ref{alg:curriculum_transfer_framework} presents our approach. The first step entails generating the LF target task from the HF target task. To obtain the task parameters for the LF task, we pass the HF target task parameters ($P^{HF_U}$) through the inverse of the affine mapping functions $ f^{-1}(P^{HF_U}) \: \forall f \in \mathcal{F}$ followed by {\tt Generate\_Env} (line 1), to obtain the parameters for the target task in the LF environment. The two requirements for obtaining a corresponding mapping are as follows:
\begin{itemize}
    \item 
    Each task parameter in the HF environment ($P^{HF}$) needs a corresponding task parameter in the LF environment ($P^{LF}$).
    $\forall \: P^{HF}_i\!\in\!P^{HF} \: \exists \: P^{LF}_i$ s.t. $f_i(P^{LF}_i) = P^{HF}_i$.
    Varying these task parameters yields different tasks that are sequenced to form a curriculum.
    \item The final task in the HF environment must be mapped to the final task in the LF environment. 
    $\forall f \in \mathcal{F} \:\exists \: f(P^{LF_U}) = P^{HF_U}$
    Thus, we can guarantee that each source task obtained through the curriculum in the LF can be mapped to a corresponding task for the HF environment.
\end{itemize}

The actions in the HF can be complex, but each action can be simplified to an LF action that need not be correspondingly equivalent. Our approach does not assume an equivalency between the state or the action spaces between the LF and the HF environments, but only on the two requirements listed above. To generate and sequence the source tasks in the LF environment, we compared two approaches. The first approach, called Handcrafted Curriculum Transfer (HC), involves a human expert deciding the parametric variables $P^{HC}$ for the tasks for the curriculum (line 3). The second approach, Automated Curriculum Transfer (AC), automatically generates and optimizes a sequence of source tasks from the parametric variables $P^{AC}$ for the agent to learn the final task with the fewest number of episodes (line 5).

\vspace{-0.5em}
\subsection{Handcrafted and Automated Curriculum Transfer Generation}




To optimize curricula through Handcrafted Curriculum Transfer ({\tt Generate\_HC}) (HC), a human expert determines the parametric variables $P^{HC}$ and the task sequence for the source tasks in the curriculum. 
Whereas, for optimizing curriculum using the Automated (AC) procedure ({\tt Generate\_AC}), we use the approach given in Algorithm \ref{alg:automated_curr_generation}. We start from an empty sequence of source task parameters $P_W$. The algorithm calls a parameterizing function ({\tt Init\_Src}) that assigns random values to the parametric variables for the first source task $P^{LF_1}$ from the range of values $P^{AC}$ can attain while simultaneously initializing an RL agent ({\tt Init\_Agent}). Based on $P^{LF_1}$, the algorithm generates the first task for the agent, $M^{LF}_1$, using the function ({\tt Generate\_Env}). The agent attempts learning this source task $({\tt Learn})$ with the initial policy $\pi_{1,w,init}$, until the \textit{stopping criterion} is met. The \textit{stopping criterion} determines if the agent's goal rate ($\delta$) is $\! \geq \! \delta_G \!$ in the last $ s $ episodes (Algorithm \ref{alg:curriculum_transfer_framework} line 12). Failure to meet the \textit{stopping criterion} implies that the agent has reached maximum permitted episodes (termed \textit{budget} ($b$)) signifying $\delta \! < \!\delta_G$. The value for $\delta_G$ is set at $0.85$ for our experiments.






\begin{algorithm}[b]
\small
\caption{{\tt ACuTE}($N\!,\!W,\!U,\!P^{HF_U},\!f,\!s,\!b$)}
\label{alg:curriculum_transfer_framework}
\raggedright \textbf{Output}: HF target task policy: $\pi^{HF}_U$ \\
\textbf{Algorithm:}
\begin{algorithmic}[1] 
\STATE $M^{LF}_U \xleftarrow[]{}$ {\tt Generate\_Env(}$f^{-1}(P^{HF_U})$)
\IF{{\tt curriculum = HC}}
    \STATE $C^{LF}= ${\tt Generate\_HC}$(M^{LF}_U, U)$ \\
\ELSIF{{\tt curriculum = AC}}
    \STATE $C^{LF}$={\tt Generate\_AC}$(M^{LF}_U, U, N, W, seeds)$\\
\ENDIF

\STATE $\pi^{HF}_0 \xleftarrow[]{} \emptyset$
\FOR{$u \in U$}
    \STATE $M^{HF}_u \xleftarrow[]{} f(C^{LF}_u)$
    \WHILE{\textit{episode} $< b$}
        \STATE $\pi^{HF}_u$ = {\tt Learn}$(M^{HF}_u, \pi^{HF}_{u-1})$
        \IF{$\mathbb{E}[\pi^{HF}_u[episode\!-\!s\!:] ] \geq \delta_G$}
            \STATE \textbf{break}
        \ENDIF
    \ENDWHILE
\ENDFOR

\STATE \textbf{return} $\pi^{HF}_U$

\end{algorithmic}
\end{algorithm}
\hfill

\vspace{-1em}

In AC, the first task of the curriculum is randomly initialized $N$ times and learned until \textit{stopping criterion} is met. The algorithm then finds the $W$ most promising solutions (${\tt Best\_Candidates}$), based on fewest interactions to meet the \textit{stopping criterion}. To optimize the sequence of the curriculum, we use beam search \cite{lowerre1976harpy}. Beam search is a greedy search algorithm that uses a breadth-first search approach to formulate a tree. At each level of the tree, the algorithm sorts all the ($N \times W$) successors of the tree ($N$ successor tasks for each task of $W$) at the current level in increasing order of number of episodes required to learn the task $M^{LF}_{u,w,n}$. Then, it selects $(W)$ number of best tasks at each level, given by fewest interactions to reach \textit{stopping criterion}, and performs the same step until the tree reaches the desired number of levels $(U)$.

\begin{algorithm}[t]
\caption{{\tt Generate\_AC($N,W,U,M^{LF}_U, seeds$)}}
\label{alg:automated_curr_generation}
\raggedright \textbf{Output}: LF Curriculum Parameters: $P_W$\\
\textbf{Placeholder Initialization}:
Timesteps: $T_1,\ldots, T_U \! \leftarrow \!\emptyset$ \\ LF task params for all tasks at each beam level $\xi_1,\ldots, \xi_U \leftarrow \emptyset$\\
LF task params at each width and level $\xi_{1,W}, \xi_{2,W}, \ldots, \xi_{U,W} \leftarrow \emptyset$\\
LF curriculum params ${P_W} \leftarrow \emptyset$ \\LF task policies for all tasks at each beam level: $\Pi_1, \ldots, \Pi_U \leftarrow \emptyset$\\
LF task policies for each width and level: $\Pi_{1,W}, \ldots, \Pi_{U,W} \leftarrow \emptyset$\\
\textbf{Algorithm:}
\begin{algorithmic}[1] 
\small
\FOR{$u \in U$}
    \FOR{$w \in W$}
        \FOR{$n \in N$}
            \IF{$u = 1$}
                \STATE $P^{LF_u} \leftarrow$ {\tt Init\_Src}$(M^{LF}_U)$
                \STATE $\pi_{1,w,init} = $ {\tt Init\_Agent}$(seeds)$
            \ELSE 
                \STATE $P^{LF_u} \leftarrow$ {\tt Init\_Inter}$(\!\xi_{u-1,W}[w],\! M^{LF}_U)$
                \STATE $\pi_{u,w, init} = $ {\tt Load\_Agent}$(\Pi_{u-1,W}[w])$ 
            \ENDIF
            \STATE $\xi_u\;[w,n] \leftarrow P^{LF_u}$
            \STATE $M^{LF}_{u,w,n} = $ {\tt Generate\_Env}$(P^{LF_u})$
            \STATE $(t_{u,w,n}, \pi_{u,w,,n,fin}) = $ {\tt Learn}${(M^{LF}_{u,w,n}, \pi_{u,w,n,init}})$
            \STATE $T_u\;[w,n] \leftarrow t_{u,w,n} \:, \: \Pi_u\;[w,n] \leftarrow \pi_{u,w,n,fin} $
        \ENDFOR
    \ENDFOR
    \STATE $T_{u,\!W}, \Pi_{u,\!W}, \xi_{u,\!W} $= {\tt Best\_Candidates}$(T_u, \!\Pi_u,\! W, \!\xi_u)$
\ENDFOR
\STATE $P_W \leftarrow $ {\tt Best\_LF\_Params}$(\xi_{1,W}, \xi_{2,W}, \ldots, \xi_{U,W})$
\STATE \textbf{return} $P_W$
\end{algorithmic}
\end{algorithm}

Now, using the parametric variables $P^{LF_1}$ for each task in the beam ($w\! \in\! W$), the algorithm generates parametric variables $P^{LF_2}$ (${\tt Init\_Inter}$) for the next task $M^{LF}_2$ in the curriculum. This is done by choosing a goal $P_G$ not encountered by the agent until the current level $u$ in the beam $w$, and randomly initializing parametric variables $\geq$ the minimum required to accomplish this goal. The agent then attempts learning $M^{LF}_2$ with the final policy of the previous source task in the beam $\pi_{1,w,fin}$ (${\tt Load\_Agent}$). The task terminates when the agent meets the \textit{stopping criterion}. The algorithm finds the $W$ most promising solutions, given by fewest interactions to reach \textit{stopping criterion} (${\tt Best\_Candidates}$) and carries out this procedure iteratively, until the final target task $M^{LF}_{U}$ is learned. 
The parameters of the curriculum with the lowest number of episodes to reach the \textit{stopping criterion} is selected as the most promising solution (${\tt Best\_LF\_Params}$) $P_W$ for learning the target task. The curriculum generation procedure requires the length of the curriculum $U$ to be $>=$ the number of goals attainable. This ensures all the goals available ($P_G$) are encountered by the agent in the curriculum. 


\subsubsection{Task Sequencing and Learning in HF}

Once the LF curriculum parametric variables ($P_W$) are obtained, they are passed through the set of mapping functions ($\mathcal{F}$) to attain the task parameters in HF, generating the curriculum source tasks from these parameters. The agent attempts learning the first source task $M^{HF}_1$ with an initial policy $\pi^{HF}_{1,init}$. The task terminates when the agent meets the \textit{stopping criterion}, generating the final policy $\pi^{HF}_{1,fin}$. This learned policy is used as an initial policy for the next source task $M^{HF}_2 $ in the curriculum. This procedure is carried out iteratively, culminating at the HF target task, returning $\pi^{HF}_U$ (Algorithm~\ref{alg:curriculum_transfer_framework}, line 8-15).

\section{Experimental Results}

\begin{figure*}
    \begin{center}
        \begin{tabular}{ c c c }
            {\includegraphics[width=0.33\linewidth,height=0.23\textwidth]{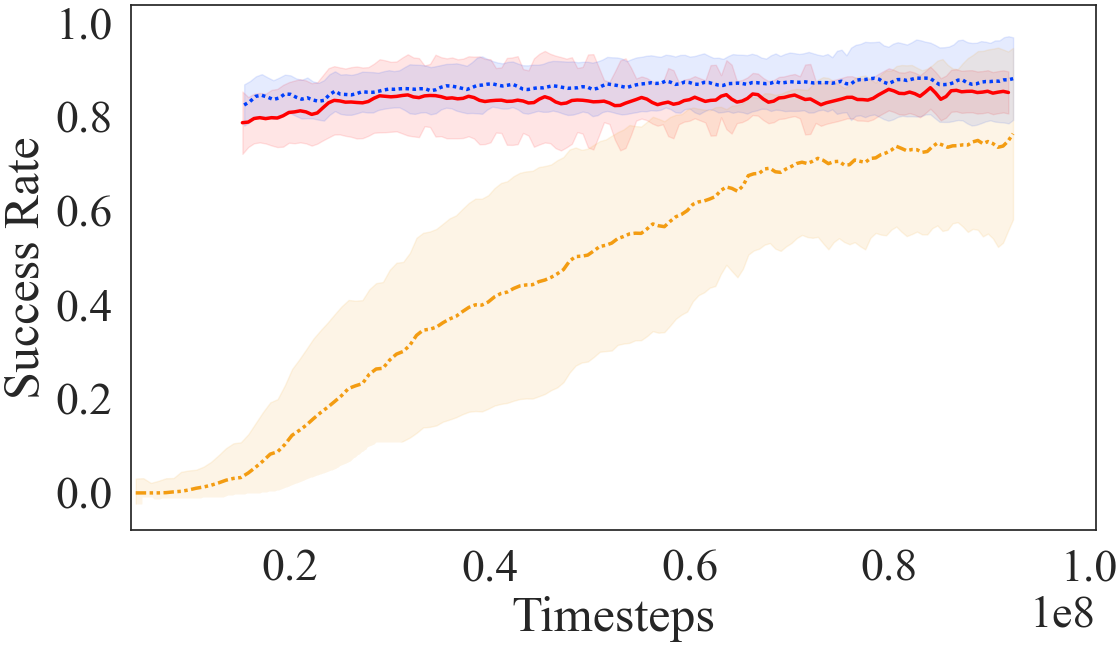}} &
            {\includegraphics[width=0.31\linewidth,height=0.23\textwidth]{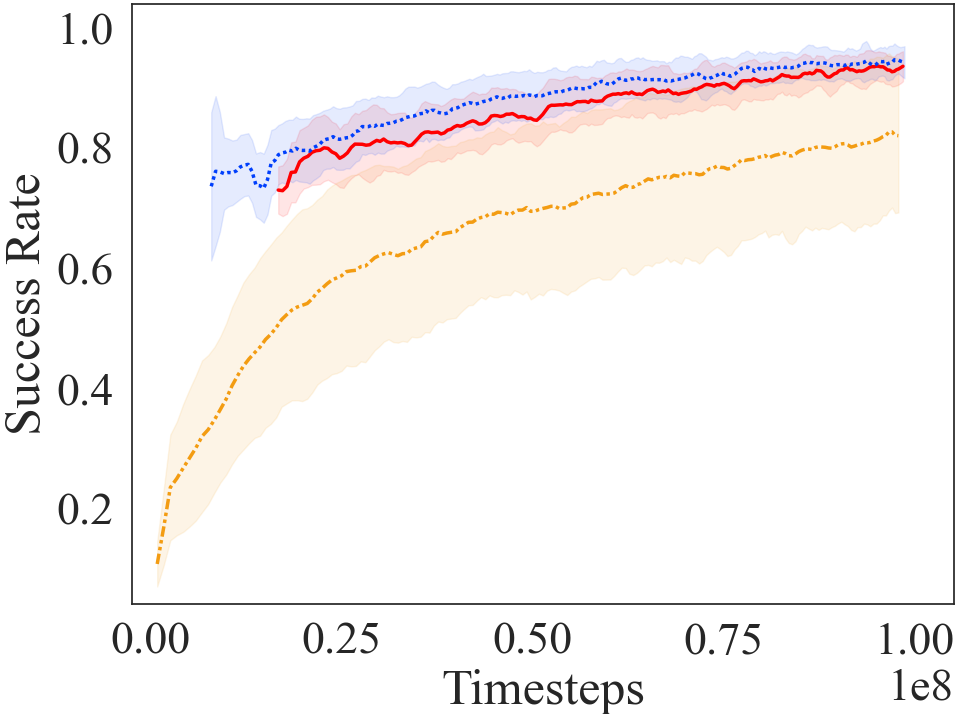}}&
            {\includegraphics[width=0.31\linewidth,height=0.23\textwidth]{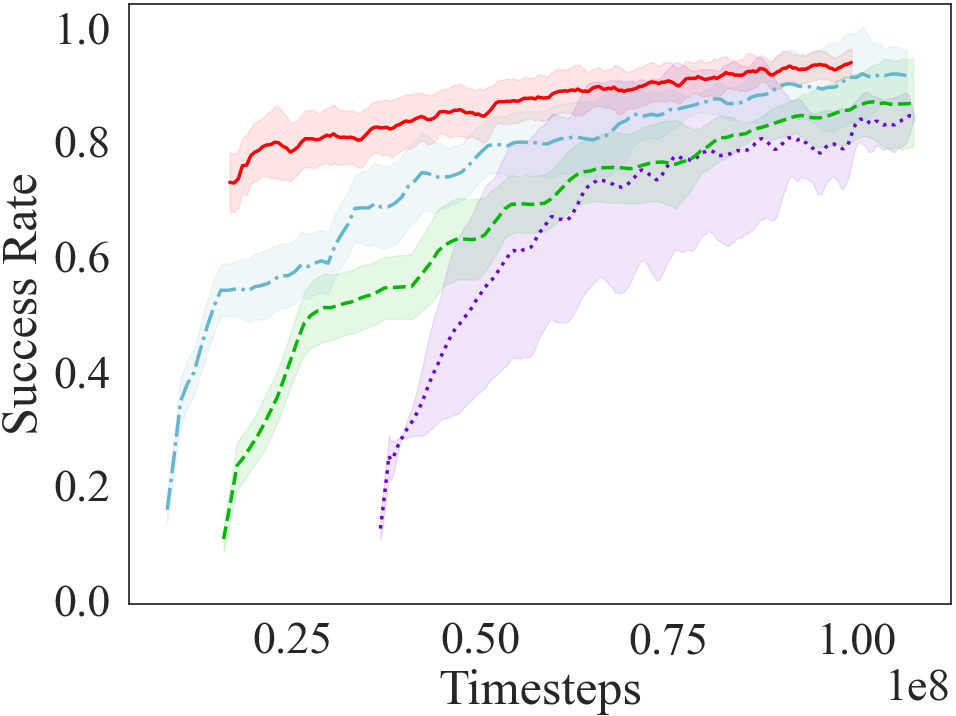}}
            \\
            {\small{\textbf{(a) LF - gridworld}}} &
            {\small{\textbf{(b) HF - Crafter-Turtlebot}}} &
            {\small{\textbf{(c) Crafter-Turtlebot Baselines}}}
            \\
            {\includegraphics[width=0.33\linewidth,height=0.23\textwidth]{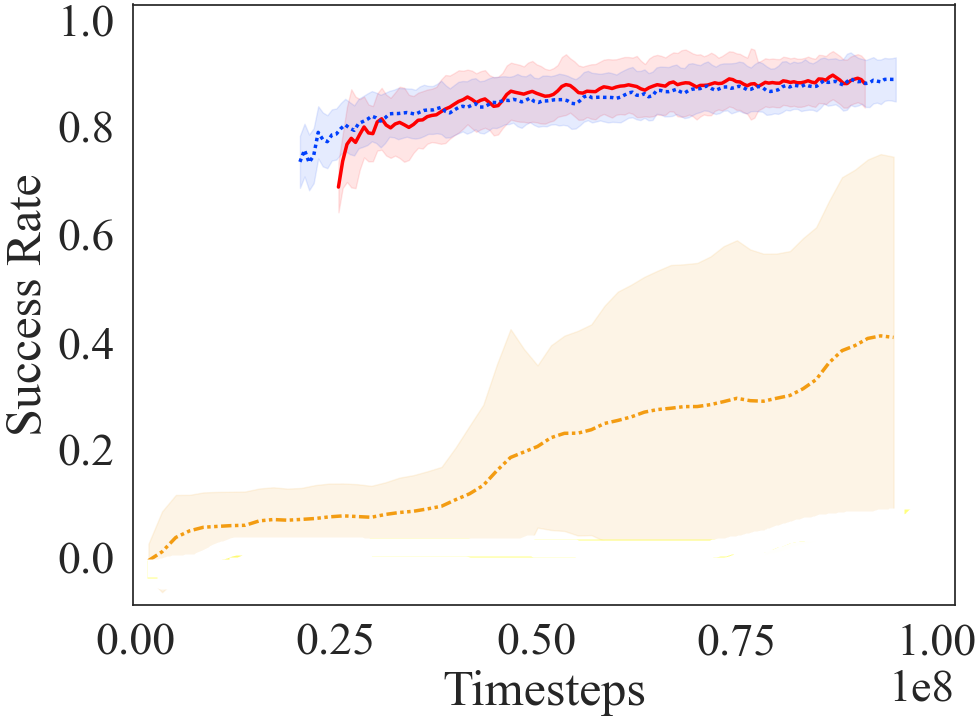}} &
            {\includegraphics[width=0.31\linewidth,height=0.23\textwidth]{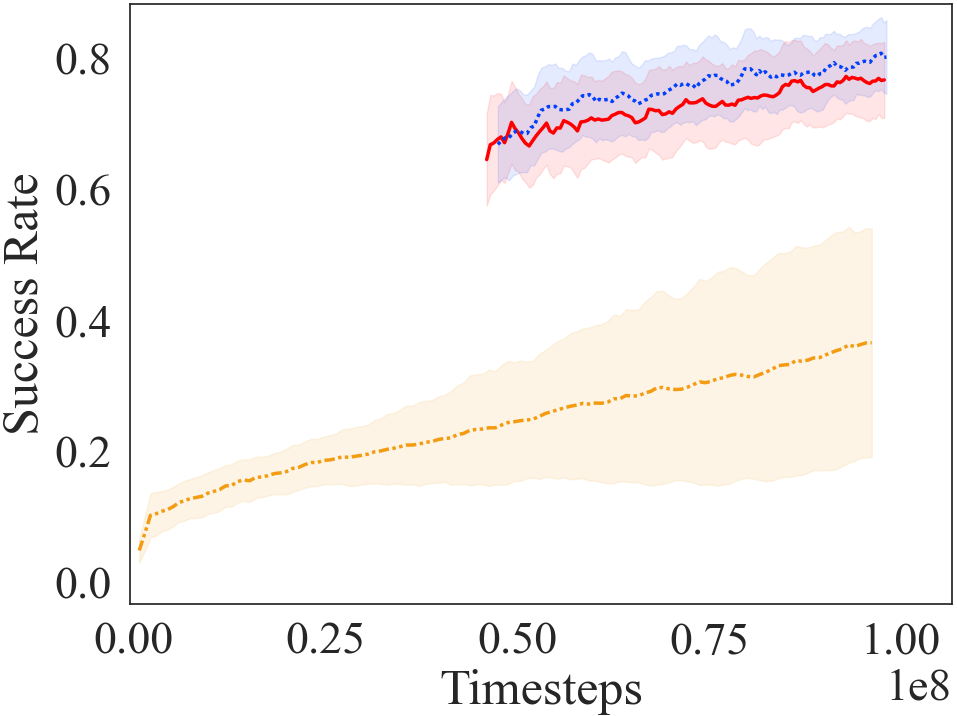}}&
            {\includegraphics[width=0.31\linewidth,height=0.23\textwidth]{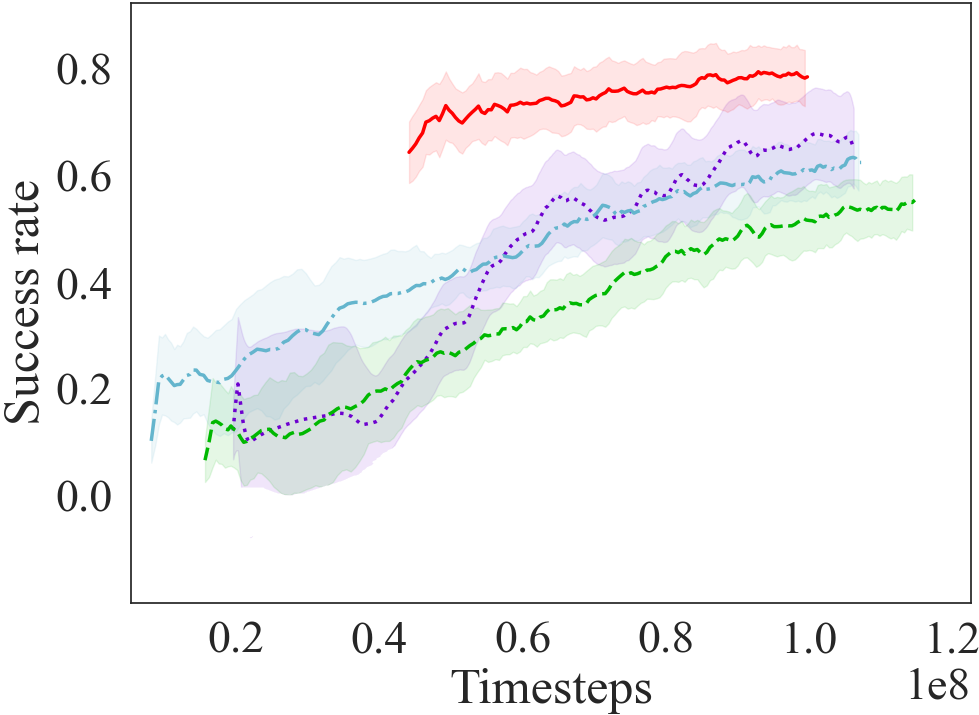}}
            \\
            {\small{\textbf{(d) LF w/ Fire - gridworld}}} &
            {\small{\textbf{(e) HF w/ Fire - Crafter-Turtlebot}}} &
            {\small{\textbf{(f) Crafter-Turtlebot w/ Fire - Baselines}}}  
        \end{tabular}
    \end{center}

\hspace{10mm}
	\begin{minipage}{0.8\textwidth} \includegraphics[width=\linewidth,height=0.08\textwidth]{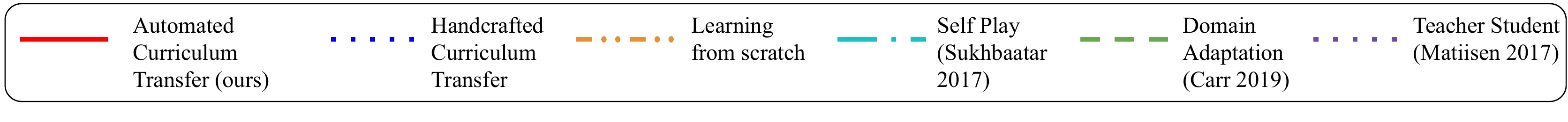}
	\end{minipage}
\vspace{-1em}
\caption{\small{Learning Curves for low fidelity (LF) and high fidelity (HF) environments with and without Fire.}} \label{fig:learning_curves}
\vspace{-1em}
\end{figure*}

We aim to answer the following questions: (1) Does the automated curriculum transfer yield sample efficient learning? (2) Does it scale to environments that are too difficult to learn from scratch? (3) Is the curriculum transfer framework independent of the RL algorithm used to generate the curriculum? (4) Can we perform a Sim2Real transfer to a physical robot? (5) Can the curriculum transfer framework yield successful convergence with imperfect (e.g., noisy) mappings between the HF and LF environments?
\footnote{Code available at: \url{https://github.com/tufts-ai-robotics-group/ACuTE}} 

To answer the first question, we evaluate our curriculum transfer method on grid-world as low-fidelity (LF) and Crafter-TurtleBot as the high-fidelity (HF) environments. In the LF environment, the agent can move 1 cell forward if the cell ahead is clear or rotate $\pi/2$ clockwise or counter-clockwise. In the target task, the agent receives a reward of $+1\!\times\!10^3$ upon crafting a stone axe, and $-1$ reward for all other steps. In the source tasks of the curriculum, the agent also receives $+50$ reward for breaking an item that is needed for crafting. This reward shaping is absent in the final target task. The agent's sensor emits a beam at incremental angles of $\pi/4$ to determine the closest object in the angle of the beam (i.e., the agents received a local-view of its environment). Two additional sensors provide information on the amount of wood and stones in the agent's inventory (See Appendix Section A.1 for further details).

The HF environment, Crafter-TurtleBot, is structurally similar to the grid-world domain, but differs in that objects are placed in continuous locations to more closely model the real-world. The agent is a TurtleBot, rendered in PyBullet \cite{coumans2019}. An example of the LF $\leftrightarrow$ HF mapping between the tasks in LF and HF environments is shown in Fig \ref{fig:one-to-one_example}. Here, the task mapping is demonstrated on an intermediate task of the curriculum, whose goal is to break a rock and craft a stone axe at the crafting table. The task mapping function ensures the LF and HF tasks have the same number and types of objects in the environments and the inventory. The mapping considers increased navigable area in the HF and does not assume the positions of the objects are preserved. In each episode, objects are positioned randomly within the boundaries of the environment. Refer to Appendix A.6 for details on mapping function set $\mathcal{F}$.

In the HF environment, the agent's navigation actions are moving forward 0.25 units and rotating by $\frac{\pi}{9}$ radians. The break and craft actions and the reward shaping in source tasks is identical to the LF environment. The HF agent's sensor is similar to the LF agent's sensor; however, it emits beams at incremental angles of $\frac{\pi}{10}$, accounting for the large state space of the location of objects.

To evaluate the performance of curriculum transfer, we used the \textit{jumpstart} \cite{lazaric2012transfer, foglino2019optimization} metric. \textit{Jumpstart} evaluates the  performance increase over $D$ episodes after transfer from a source task as compared to a baseline approach. \textit{Jumpstart} is defined as:
\vspace{-1mm}
\[ \eta_j := \frac{1}{D} \sum_{i=1}^D (G^i_{M^c_f}  - G^i_{M^b_f})\]
\vspace{-1.5mm}
\noindent where $G^i_{M^c_f}$ and $G^i_{M^b_f}$ are the returns obtained during episode $i$ in task $M^c_f$ (learning through automated curriculum transfer) and the baseline task $M^b_f$ respectively. Another metric we used is the \textit{time to threshold} metric \cite{taylor2009transfer, narvekar2020curriculum}, which computes how faster an agent can learn a policy that achieves expected return $G \geq \delta$ on the target task if it transfers knowledge, as opposed to learning from another approach, where $\delta$ is desired performance threshold.

\vspace{-0.45em}
\subsection{Curriculum Generation in High Fidelity} \label{subsection:curricula_generation_results}

We used the algorithm presented in Algorithm \ref{alg:curriculum_transfer_framework} to generate and sequence the source tasks using the handcrafted curriculum transfer (HC) and the automated curriculum transfer (AC) approach. The navigable area in target task $M^{LF}_U$ in low-fidelity (LF) environment is a grid of $\mathbb{R}_{LF}^2 \!\xrightarrow{} [10\! \times\!10]$, as seen in Fig \ref{fig:Final_tasks}a, and the high-fidelity (HF) target task area is a continuous $\mathbb{R}_{HF}^2 \!\xrightarrow{} [4 m\! \times\! 4 m]$ space, as seen in Fig \ref{fig:Final_tasks}b. Both these environments contain 4 trees, 2 rocks and 1 crafting table, placed at random locations.

The RL algorithm was a Policy Gradient \cite{williams1992simple} network with $\epsilon$-greedy action selection for learning the optimal policy. The episode terminates when the agent successfully crafts a stone axe or exceeds the total number of timesteps permitted, which is $10^2$ in the LF environment and $6\!\times\!10^2$ in the HF environment. All experiments are averaged over 10 trials. (See Appendix Section A.2).

For generating the AC in the LF environment, we set the width of the beam search algorithm at $W\!=\!4$, and the length of the beam at $N\!=\!20$, the curriculum length was $U\! =\! 4$, and \textit{budget} $\!b \! = \! 5\times10^3$. We obtained the parameters after performing a heuristic grid search over the space of the search algorithm. We evaluated our curriculum transfer framework with four other baselines. \textit{Learning from scratch} trains the final HF target task without any curriculum. We also adopted three approaches from the literature designed for speeding-up RL agents: Asymmetric Self-play \cite{sukhbaatar2018intrinsic}, Teacher-Student Curriculum learning~\cite{DBLP:journals/tnn/MatiisenOCS20} and Domain Adaptation for RL \cite{carr2019domain}. The first two baselines do not make use of the LF environment while the third uses the LF environment as the source domain. All baseline approaches involve reward shaping similar to the source task of the automated curriculum transfer approach.

The results in Figs \ref{fig:learning_curves}a and \ref{fig:learning_curves}b show that the AC approach results in a substantial improvement in learning speed, and is comparable to the curriculum proposed by a human expert. Furthermore, as shown in Fig~\ref{fig:learning_curves}f, the curriculum transfer method outperforms the three baseline approaches\footnote{Refer Appendix Section A.5 for learning curves for reward} in terms of learning speed. The learning curve for our curriculum transfer approaches has an offset on the x-axis to account for the time steps used to go through the curriculum before moving on to the target task, signifying \textit{strong transfer} \cite{taylor2009transfer}. The other three baseline approaches perform better than learning from scratch, but do not outperform the curriculum transfer approach. Self-Play requires training a goal-proposing agent, which contributes to the sunk cost for learning. Whereas, Domain Adaptation relies on the similarity between the tasks. Teacher-Student curriculum learning requires defining the source tasks of the curriculum beforehand, and the teacher attempts to optimize the order of the tasks. All baselines involve interactions in the costly high-fidelity domain for generating/optimizing the curriculum, which proves to be costly. This sunk cost has been accounted in the learning curves by having an offset on the x-axis. See Appendix Section A.3 for details on our adaptation and tuning of these three baselines.

Table \ref{table:jumpstart_metric} compares the \textit{jumpstart} values for AC with the baselines. The higher \textit{jumpstart}, the better performance of the curriculum. AC achieves high positive \textit{jumpstart} and \textit{time to threshold} performance in comparison to baseline approaches, denoting improved performance and quicker learning (magnitude $> \!10^7$ timesteps).

\subsection{Results with Added Complexity} 
To answer the second question, we evaluated our framework in a situation where the target task in the HF environment is too difficult to learn from scratch. To do this, both the LF and HF environments were modified by adding a new type of object, ``fire'', such that when the agent comes in contact with it, the episode terminates instantly with a reward of $\!-1\!\times\!10^3$. 

\begin{table}[t]
\small
\begin{tabular}{|p{0.09\linewidth}|p{0.38\linewidth}|p{0.16\linewidth}|p{0.16\linewidth}|}
\hline
 Env & Methods & $\Delta$Jumpstart (Return) & $\Delta$Time-to-threshold\\
 \hline
 \multirow{3}{*}{HF}&AC $\xrightarrow[]{}$ Learning from scratch &304 $\pm$ 242& 5.4 $\times 10^7$ \\
    &AC $\xrightarrow[]{}$ Carr~\emph{et.~al.}& 231$\pm$160&3.5 $\times 10^7$\\
    &AC $\xrightarrow[]{}$ Sukhbaatar \emph{et.~al.}& 85$\pm$130&2.1 $\times 10^7$\\
    &AC $\xrightarrow[]{}$ Matiisen \emph{et.~al.}& 568 $\pm$205&6.3 $\times 10^7$\\    
\hline
 \multirow{3}{*}{\shortstack[l]{\\HF\\w/\\Fire}}&
 AC $\xrightarrow[]{}$ Learning from scratch& 628$\pm$284&1.02 $\times 10^8$\\
    &AC $\xrightarrow[]{}$ Carr~\emph{et.~al.}& 519$\pm$126&8.4 $\times 10^7$\\
    &AC $\xrightarrow[]{}$ Sukhbaatar \emph{et.~al.}& 288$\pm$87&7.8 $\times 10^7$\\
    &AC $\xrightarrow[]{}$ Matiisen \emph{et.~al.}& 346$\pm$121&4.6 $\times 10^7$\\    
 \hline
\end{tabular}
\caption{\small{Table comparing \textit{jumpstart} (mean$\pm$SD), and \textit{time to threshold} for learning the final target task. Here, HF and AC refer to high-fidelity and automated curriculum transfer respectively. Time-to-threshold measured in timesteps.}}
\label{table:jumpstart_metric}
\vspace{-3em}
\end{table}

\begin{figure}[b]
\vspace{-1em}
\includegraphics[width=0.7\linewidth,height=0.45\linewidth]{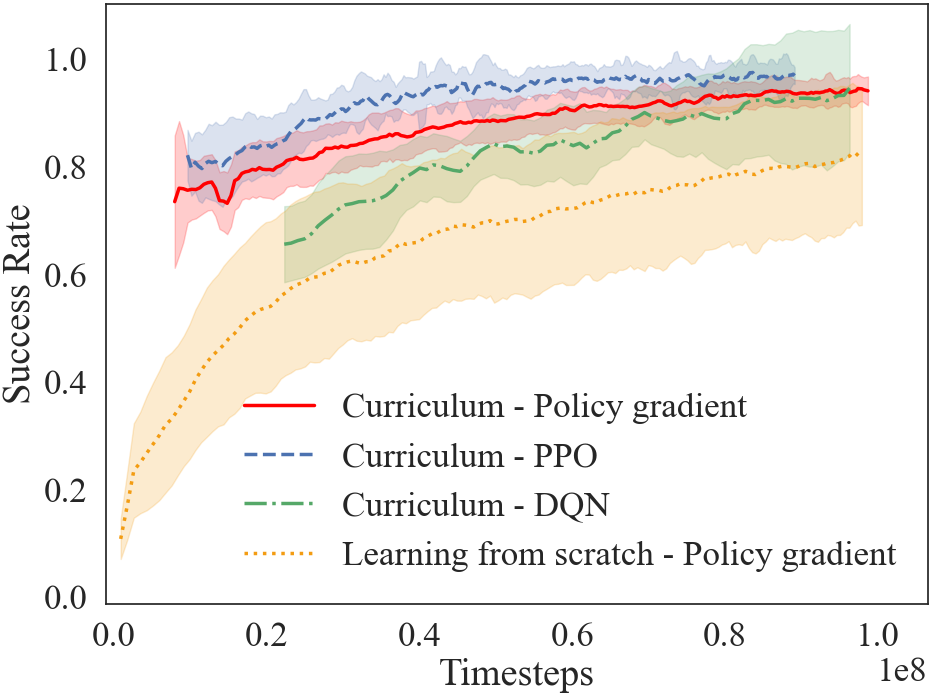}
\caption{\small{Learning HF target task through different RL algorithms.}}
\label{fig:A2C}
\end{figure}

The parameters listed in Section \ref{subsection:curricula_generation_results} are used to optimize AC in the LF environment. From learning curves in Figs. \ref{fig:learning_curves}d, \ref{fig:learning_curves}e, \ref{fig:learning_curves}f and Table \ref{table:jumpstart_metric}, we observe our curriculum transfer approach, AC, consistently achieving higher average reward, better \textit{jumpstart} and \textit{time to threshold} performance compared to baselines. Learning from scratch fails to converge to a successful policy in $10^8$ interactions, while the other baseline achieves marginally better performance than learning from scratch. Table~\ref{table:jumpstart_metric} summarizes the jumpstart and the time-to-threshold between the approaches. Our approach still achieves a high jumpstart, and converges much quicker than other approaches. Through this experiment, we see our AC approach extrapolates to challenging environments, producing quicker and efficient convergence. Refer Appendix section A.1.1 for trends observed in different runs in our AC approach.
\vspace{-0.6em}
\subsection{Results with Different RL Algorithms}

To answer the third question, we conducted experiments by making the HF learning algorithm different (PPO~\cite{DBLP:journals/corr/SchulmanWDRK17} and DQN~\cite{mnih_human-level_2015}) from the RL algorithm used for generating the curriculum in the LF environment (Policy gradient). Fig \ref{fig:A2C} shows the result of this test  (See Appendix A.4 for hyperparameters). In all the cases, learning through a curriculum is quicker and more efficient than learning from scratch. Here, we do not intend to find the best RL algorithm to solve the task, but demonstrate that actor-critic networks, policy gradients and value function based approaches learn the HF target task through curriculum, irrespective of the RL algorithm used to optimize the LF curriculum.

\begin{figure*}
    \begin{center}
        \begin{tabular}{ c c c }
            \label{fig:learning_curves_noisy}

            {\includegraphics[width=0.33\linewidth,height=0.23\textwidth]{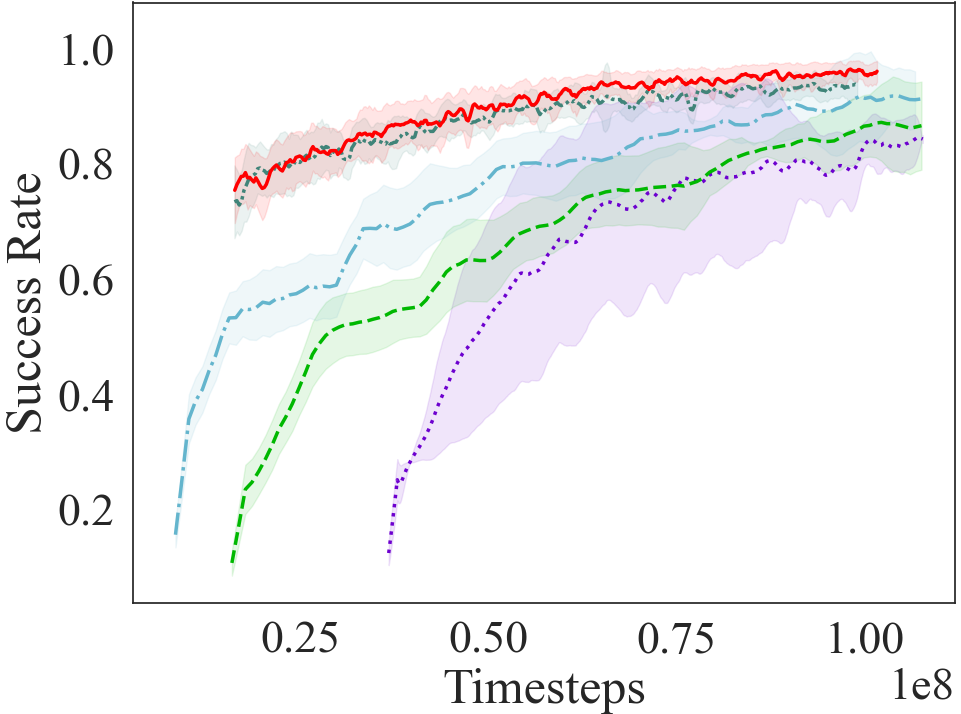}} &
            {\includegraphics[width=0.31\linewidth,height=0.23\textwidth]{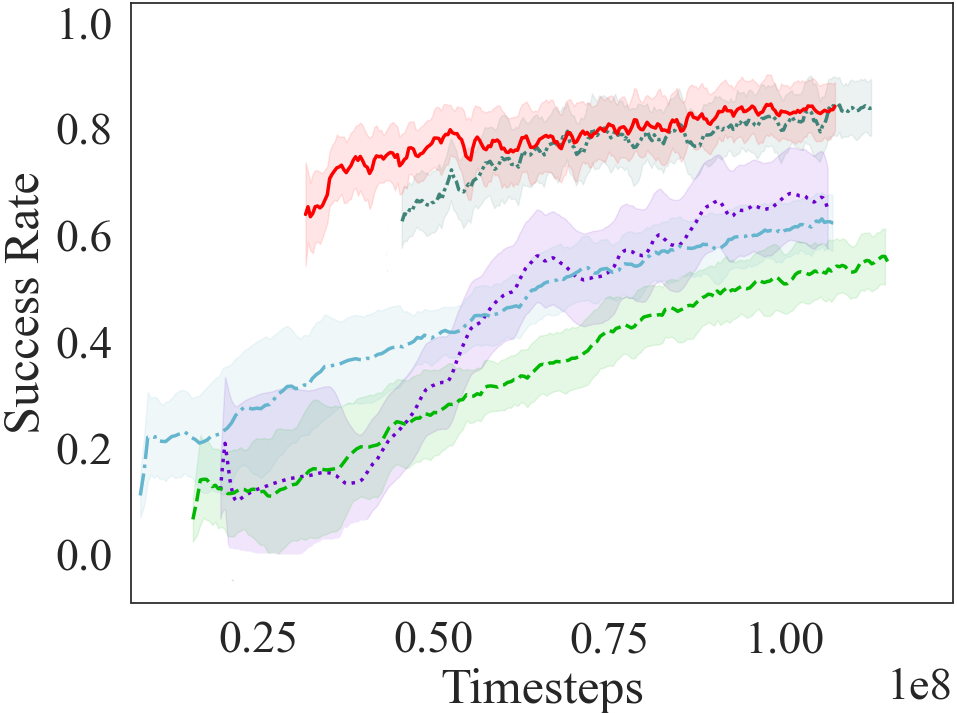}}&
            {\includegraphics[width=0.15\linewidth,height=0.23\textwidth]{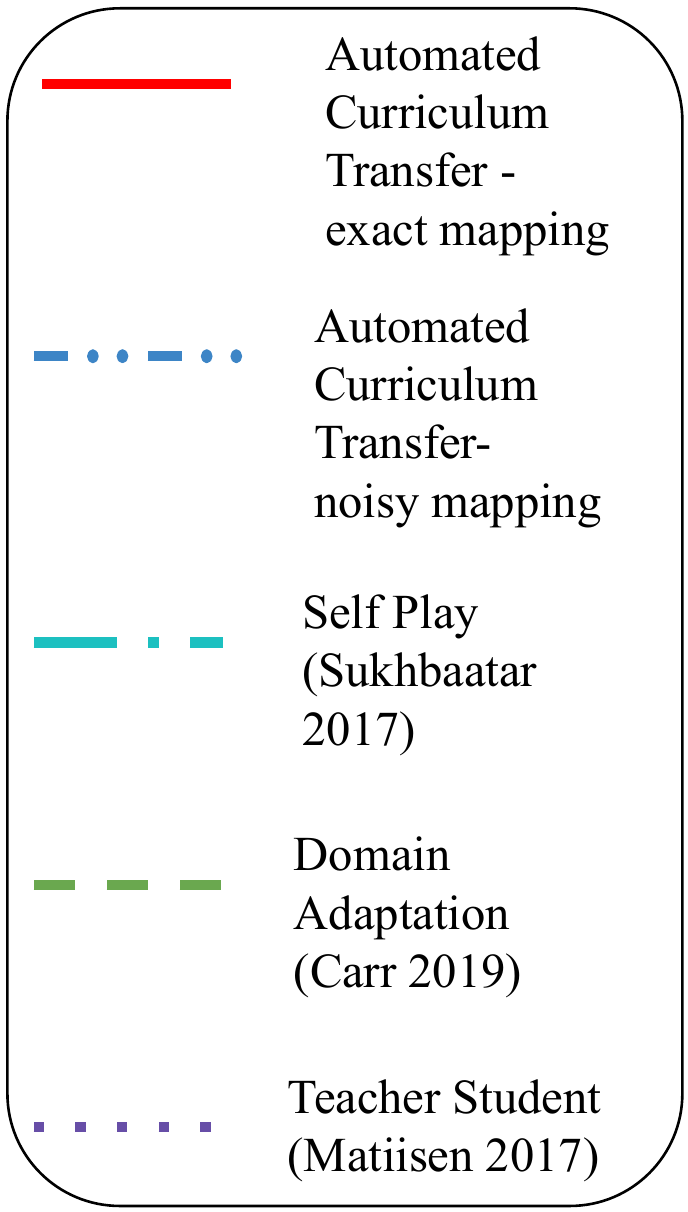}}
            \\
            {\small{\textbf{(a) Noisy mappings - HF}}} &
            {\small{\textbf{(b) Noisy mappings - HF w/ Fire}}} &
            {}
        \end{tabular}
        \caption{\small{Comparison of Learning Curves for high fidelity (HF) environments generated using noisy mappings.}} \label{fig:learning_curves_noisy}
    \end{center}
\end{figure*}


\subsection{Noisy Mappings}

In the above sections, we evaluated the curriculum transfer schema on accurate mappings between the two environments. In certain partially observable environments, it might not be possible to obtain such an accurate mapping between the two environments. To demonstrate the efficacy of our approach in imperfect mappings, we evaluate the experiments by incorporating noise in the mapping function. The noisy mapping function involves a multivariate noise over the range of the parametric variables. While obtaining the noisy parameters, we do not incorporate any noise in the parameter for the goal condition, as we can safely assume that the goals have been mapped accurately. The noisy mappings are given by:
\[P_{noisy} = P_{exact} + \mathcal{N}(0, \Sigma) \] 
where $\Sigma = \begin{bmatrix}
\sigma_1 & 0 & 0 & \ldots\\
0 & \sigma_2 & 0 & \ldots\\
\ldots & \ldots & \ldots & \ldots\\
\ldots & \ldots & \ldots & \sigma_n
\end{bmatrix}_{n \times n}$is the covariance matrix, which is symmetric and positive semi-definite, and $\sigma_i = (max(P_i) - min(P_i))/6$, covers the entire range of the parametric variable in six standard deviations. We verify whether the noisy parameters meet the minimum requirements of reaching the goal for the sub-task. If the requirements are not met, we generate a new set of noisy parameters until the requirements are met. 

The learning curves for the automated curriculum transfer generated through noisy parameters are shown in Fig~\ref{fig:learning_curves_noisy}. We compare its performance with learning curves for automated curriculum transfer generated through exact mappings, and with other baseline approaches present in the literature. Even with noisy mappings, the automated curriculum transfer outperforms other curriculum approaches, and performs comparable to the automated curriculum transfer with exact mappings. On the complicated task (HF with Fire), the automated curriculum transfer with noisy mappings takes longer to converge on the source tasks of the curriculum, yet it achieves a significant performance advantage over other baselines. 

\subsection{Runtime Comparison}

\begin{figure}[t]
\includegraphics[width=0.8\linewidth,height=0.45\linewidth]{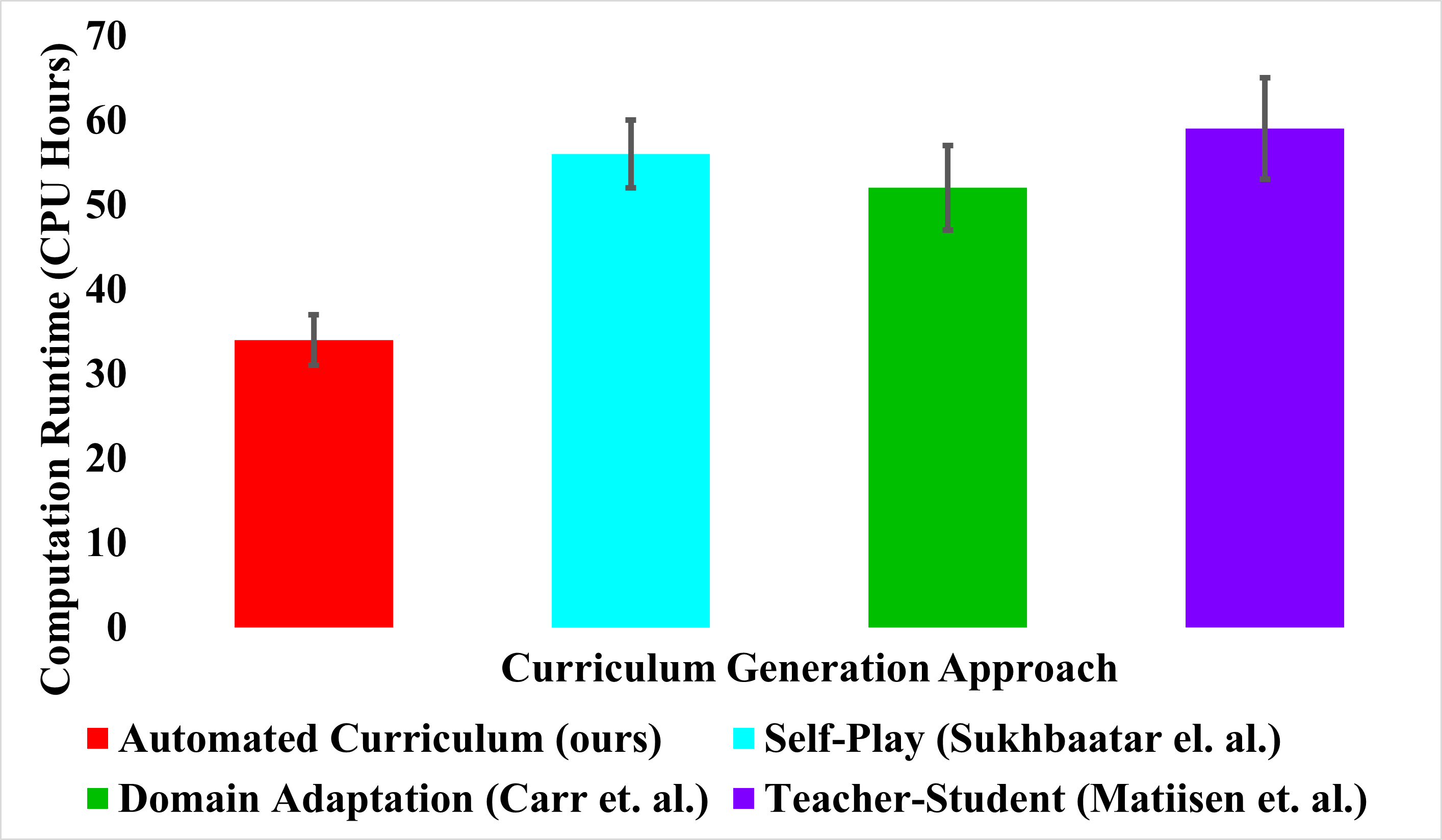}
\vspace{-1em}
\caption{\small{Runtime comparison of different approaches.}}
\label{fig:runtime}
\vspace{-1em}
\end{figure}

Since our approach relies on the low-fidelity environment for curriculum generation, its sunk cost in computational runtime is significantly lesser than baseline approaches, in which curriculum generation requires extensive interactions in the costly high-fidelity environment. Fig~\ref{fig:runtime} compares the computational runtime (in CPU Hours) required to run one trial (with each episode having a maximum of $6\times 10^2$ timesteps) of the high-fidelity task until the \textit{stopping criterion} is met. The experiments were conducted using a 64-bit Linux Machine, having Intel(R) Core(TM) i9-9940X CPU @ 3.30GHz processor and 126GB RAM memory. The sunk cost of our automated curriculum transfer approach involves the interactions required to optimize the curriculum in the LF environment, and the interactions required to learn the source tasks in the HF environment.

\section{Transfer to a Physical Robot}\label{appendix:transfer_to_robot}

To answer the fourth question and demonstrate the efficacy of this approach in the real-world, we performed a Sim2Real transfer on a TurtleBot after the simulated robot had learned the target task through the curriculum.
We demonstrate object breaking and item retrieval, followed by crafting, through a TurtleBot which reads fiducials scattered throughout the environment.

We made use of the TurtleBot~2, modified to have an on-board laptop (running Ubuntu 16.04 and ROS Kinetic) which interfaces with a camera (the Intel D435i) and a LIDAR unit (the Slamtec RPLIDAR-A2). These additional sensors were used to provide more stable odometry via the Canonical Scan Matcher \cite{censi2008icp},
as the default odometry stack was found to have too much drift to be relied upon. The full platform is visible in Fig \ref{fig:Final_tasks}c.
Making use of the modified odometry stack, movement was provided as actions that attempt movement in the approximate units expected by the high-fidelity environment. 

In this setting, the target task policy learned through our automated curriculum transfer approach in the high-fidelity Crafter-TurtleBot environment is transferred to run in this physical setting. No learning is taking place on the agent as the physical experimentation continues. These fiducials are QR codes corresponding to the different possible objects in the known environment: trees, rocks, and the crafting table.
When demonstrated, the agent controls the TurtleBot to navigate to the Tree locations before calling the break action, and is then visible navigating to the Rock locations before again calling the break action. The agent completes the sequence by navigating to the crafting table before calling the craft action. The policy is run without any modifications.

It is important to note that breaking and crafting actions are successful only in the event of reading the fiducial. In this way, we restrict the agent to being successful only in the cases where it has operated successfully in the physical environment, allowing the agent to proceed only in the event of a sensible Sim2Real transfer. 

\vspace{-0.5cm}
\section{Conclusion and Future Work}

We proposed a framework for automated curriculum transfer from an easy-to-solve environment to a real-world scenario. Our curriculum transfer approach generated results comparable to a curriculum generated by a human expert, exceeding the baseline performances. Our experimental evaluations show improved learning time and jumpstart performance on the final target task, even when additional challenging elements are introduced. We demonstrated ACuTE is independent of the RL algorithm used to generate the curriculum and is easily Sim2Real transferable to a physical robot setting and is also scalable to environments with inexact mappings.

An extension of our approach will be to scale the algorithm to multi-agent settings, with inter-agent curriculum transfer. 
A limitation of this work is that the task mapping is generated heuristically, a future work would involve automating the mapping generation. Future work can investigate how a LF version of the environment can be created autonomously, and providing a theoretical guarantee for the curriculum transfer approach. Second, while in this work we only transferred the schema of the curriculum, our baseline comparison with Domain Adaptation suggests that DA can be combined with curriculum transfer such that the agent can learn the tasks in the curriculum even faster. Finally, our algorithms for optimizing the curriculum in the LF environment did not make use of any data from the HF domain, and in future work, we plan to modify our framework to use interaction experience from both domains when constructing the curriculum.

\vspace{-0.5cm}

\begin{acks}
The research presented in this paper was conducted with partial support from DARPA (contract W911NF-20-2-0006) and AFRL (contract FA8750-22-C-0501).
\end{acks}

\bibliographystyle{ACM-Reference-Format}
\balance
\bibliography{aamas21}  







\end{document}